\documentclass[10pt,twocolumn,letterpaper]{article}
\usepackage[pagenumbers]{cvpr} 
\usepackage{graphicx}
\usepackage{amsmath}
\usepackage{amssymb}
\usepackage{booktabs}

\usepackage{multirow}
\usepackage{tabularx,verbatim}
\usepackage{xspace}
\usepackage{algorithm}
\usepackage{algorithmic}
\usepackage{setspace}
\usepackage{comment}
\usepackage{bbm}
\usepackage{enumitem}
\usepackage{colortbl}
\usepackage{color, xcolor} 
\usepackage[T1]{fontenc}
\usepackage{bbm}
\usepackage{listings}

\usepackage{pifont}
\newcommand{\cmark}{\ding{51}}%

\definecolor{remark}{rgb}{1,.5,0} 
\definecolor{citecolor}{rgb}{0,0.443,0.737} 
\definecolor{linkcolor}{rgb}{0.956,0.298,0.235} 
\definecolor{cyan}{rgb}{0.831,0.901,0.945}

\usepackage{amsmath}
\DeclareMathOperator*{\argmax}{arg\,max}

\usepackage[pagebackref,breaklinks,colorlinks,bookmarks=false]{hyperref}
\colorlet{dark-blue}{blue!70!black}
\colorlet{dark-green}{green!80!black}
\colorlet{dark-red}{red!80!black}
\definecolor{mypink}{RGB}{219, 48, 122}
\hypersetup{
    colorlinks=true,%
    citecolor=citecolor,%
    filecolor=citecolor,%
    linkcolor=linkcolor,%
    urlcolor=magenta
}
\usepackage{url}

\usepackage[capitalize]{cleveref}
\crefname{section}{Sec.}{Secs.}
\Crefname{section}{Section}{Sections}
\Crefname{table}{Table}{Tables}
\crefname{table}{Tab.}{Tabs.}

\usepackage[normalem]{ulem}

\def\cvprPaperID{208} 
\def\confName{CVPR}
\def\confYear{2022}

\begin{document}

\title{Novel Class Discovery in Semantic Segmentation}

\author{%
  Yuyang Zhao$^{\textcolor{mypink}{1}}$ \quad Zhun Zhong$^{\textcolor{mypink}{2}}$ \quad Nicu Sebe$^{\textcolor{mypink}{2}}$ \quad Gim Hee Lee$^{\textcolor{mypink}{1}}$ \\
  \small{$^{\textcolor{mypink}{1}}$ Department of Computer Science, National University of Singapore}  \\
  \small{$^{\textcolor{mypink}{2}}$ Department of Information Engineering and Computer Science, University of Trento}\\ 
 \small{Project: \href{https://ncdss.github.io}{https://ncdss.github.io} }
}

\maketitle

\begin{abstract}
We introduce a new setting of Novel Class Discovery in Semantic Segmentation (NCDSS), which aims at segmenting unlabeled images containing new classes given prior knowledge from a labeled set of disjoint classes. In contrast to existing approaches that look at novel class discovery in image classification, we focus on the more challenging semantic segmentation. In NCDSS, we need to distinguish the objects and background, and to handle the existence of multiple classes within an image, which increases the difficulty in using the unlabeled data. To tackle this new setting, we leverage the labeled base data and a saliency model to coarsely cluster novel classes for model training in our basic framework. Additionally, we propose the Entropy-based Uncertainty Modeling and Self-training (EUMS) framework to overcome noisy pseudo-labels, further improving the model performance on the novel classes. Our EUMS utilizes an entropy ranking technique and a dynamic reassignment to distill clean labels, thereby making full use of the noisy data via self-supervised learning. We build the NCDSS benchmark on the PASCAL-5$^i$ dataset and COCO-20$^i$ dataset. Extensive experiments demonstrate the feasibility of the basic framework (achieving an average mIoU of 49.81\% on PASCAL-5$^i$) and the effectiveness of EUMS framework (outperforming the basic framework by 9.28\% mIoU on PASCAL-5$^i$). 

\end{abstract}

\section{Introduction}
\label{sec:intro}

\begin{figure}[t]
    \centering
    \includegraphics[width=0.95\linewidth]{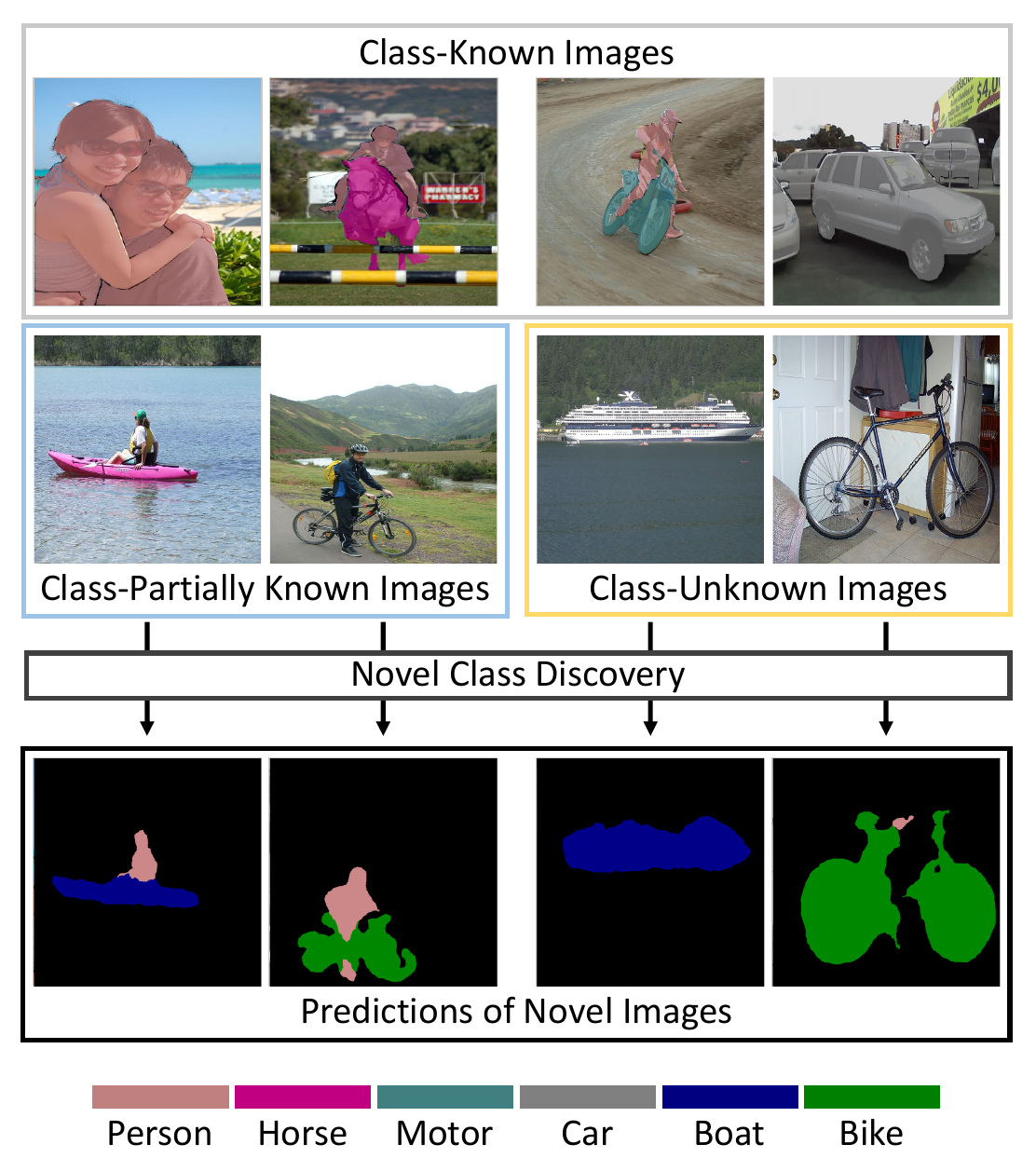}
    \vspace{-.1in}
    \caption{Illustration of Novel Class Discovery in Semantic Segmentation~(NCDSS). Given a labeled set (person, horse, motorbike and car) and a class disjoint unlabeled set (bike and boat), NCDSS aims at leveraging all the data to segment novel images.}
    \label{fig:intro}
\end{figure}

Deep Neural Networks (DNNs) have achieved significant success in semantic segmentation~\cite{deeplabv3plus, long2015fully} in recent years. However, the success of DNNs heavily relies on the large number of annotated data, which %
incurs high annotation cost, especially for semantic segmentation where pixel-wise labeling is required. Several settings are proposed to alleviate the cost, such as semi-supervised learning~(SSL), unsupervised learning~(USL), \etc.
SSL~\cite{zou2021pseudoseg, chen2021semi} utilizes a proportion of annotated data, aiming to achieve comparable performance with fully-supervised methods. Generally, SSL is based on the assumption that the labeled and unlabeled data share the same label space.  
Nevertheless, it is possible %
to have data from the unseen novel classes in real-world semantic segmentation settings. %
Such data can be easily collected but difficult to annotate. %
USL~\cite{maskcontrast, hwang2019segsort} is also %
introduced to %
mitigate the annotation cost. 
Due to the complexity of unlabeled data, USL cannot achieve satisfactory results without any prior knowledge.

In contrast to machine learning models, humans can easily discover new categories with prior knowledge. For example, a kid can easily discover sheep when he/she knows how to distinguish between horse and cow.
Based on this observation, a practical setting of Novel Class Discovery~(NCD)~\cite{Han2019learning} is introduced in the computer vision community.  
NCD aims at discovering novel classes of the unlabeled data when given a set of labeled data of base classes, where the base and novel classes are disjoint. 
In view of the practicability of NCD, previous works~\cite{Han2019learning, Fini2021uod} explore NCD in the image classification task. In this paper, 
we %
further extend NCD to semantic segmentation and introduce a new setting, called \textit{\textbf{N}ovel \textbf{C}lass \textbf{D}iscovery in \textbf{S}emantic \textbf{S}egmentation} (NCDSS), which is illustrated in Fig.~\ref{fig:intro}. Given a set of labeled data and a set of unlabeled data, the goal is to segment novel images with the prior knowledge from base foreground and background classes. 
Recently, a similar setting called weak-shot learning (WSHL)~\cite{zhou2021weak} is introduced to semantic segmentation. However, WSHL requires weak annotations for the novel data (\eg, image-level labels) and thus limiting its practicality. %
The comparison of different settings is shown in Tab.~\ref{tab:setting}.

NCDSS is %
more challenging than NCD in classification. In comparison 
with image classification that uses global information, semantic segmentation requires the model to classify the pixels of multiple foreground and background categories within one image based on the local information. 
This greatly increases the difficulty in utilizing unlabeled novel class data. 
Despite the difficulty, it is worth investigating %
NCD in semantic segmentation since it is not always possible to label all classes of semantic labels. %
To address NCDSS, we first propose \textit{a basic framework} in this paper.
Specifically,
we choose to cluster foreground features instead of the global features for more accurate clustering results. The foreground and background can be effectively separated by
estimating the saliency map from a well-trained saliency model. 
However, directly clustering the foreground pixels %
can lead to inaccurate labels
due to the multiple foreground categories within an image. %
This problem can be addressed by prior knowledge from the base data. Concretely, we use the base model to detect high confidence pixels of the base classes, and ignore them when clustering. 
Our basic framework can coarsely segment novel classes with the guidance of prior knowledge and therefore leading to a feasible solution for NCDSS.

There are three issues limiting the performance of our basic framework: 1) the estimated saliency maps are not sufficiently accurate; 2) unsupervised clustering alone cannot guarantee precise label assignments even when accurate saliency maps are given; 3) there are also circumstances that more than one novel classes may appear in one image (\eg, person and horse), and clustering can only assign one label for the novel salient part of one image. 
Consequently, the pseudo-labels generated by clustering are quite noisy (some pseudo-labels are unclean) and thus inevitably 
degrading the model performance~\cite{zhang2017understanding, arpit2017closer}. Intuitively, %
the unclean clustering labels can be discarded, and the data can be treated as additional unlabeled data to boost the performance if using only the part of clean clustering pseudo-labels can achieve comparable or better performance. 
To this end, we propose the \textit{\textbf{E}ntropy-based \textbf{U}ncertainty \textbf{M}odeling and \textbf{S}elf-training (EUMS) framework}.
EUMS uses entropy values obtained from the basic model to measure the uncertainty of novel images~\cite{intra} and splits them into clean and unclean parts. Lower entropy values indicate that the training images are assigned with more accurate pseudo-labels. For the unclean part, inspired by \cite{sohn2020fixmatch}, we use online pseudo-labels from the teacher model to train the student model in a self-supervised manner. We observe that the average accuracy of clustering pseudo-labels in the clean part are higher than that of the whole novel set. However, hard classes (\eg, potted plant) are almost completely ignored when only a 
small ratio of data is split into clean part (\eg, 0.33), which has a pernicious influence on model performance. Consequently, we start from a relatively high clean part ratio to learn a good model initialization, and then dynamically re-rank the clean data and reassign more data into the unclean part. This strategy encourages the model not to ignore any category and thus obtains further improvement. 

Our main contributions can be summarized as follows:
\begin{itemize}
\setlength{\itemsep}{5pt}
\setlength{\parsep}{0pt}
\setlength{\parskip}{0pt}

\item We introduce a new setting of Novel Class Discovery in Semantic Segmentation (NCDSS), which is a practical but underexplored problem. In addition, we propose a basic framework which utilizes base class prior knowledge to discover novel categories.

\item We propose the EUMS framework that distills accurate (clean) clustering pseudo-labels with entropy ranking and utilizing unclean data via self-training, which can effectively improve the model performance.

\item We build the NCDSS benchmark on the PASCAL-5$^i$ dataset and COCO-20$^i$ dataset. Extensive experiments on the two benchmarks verify the effectiveness of our basic and EUMS frameworks. 

\end{itemize}


\begin{table}[t]
    \centering
    \small
    \begin{tabular}{l|cc}
    \toprule
        Setting & Base Classes & Novel Classes \\
    \midrule
         SL & fully & --  \\
         SSL & semi & --  \\
         USL & unlabeled & -- \\
         WSHL & fully & weakly \\
         NCD & fully & unlabeled \\
    \bottomrule
    \end{tabular}
    \vspace{-.1in}
    \caption{Comparison of different semantic segmentation settings.}
    \label{tab:setting}
\end{table}
\vspace{-.1in}

\section{Related Work}
\label{sec:;related}

\begin{figure*}[t]
    \centering
    \includegraphics[width=0.95\linewidth]{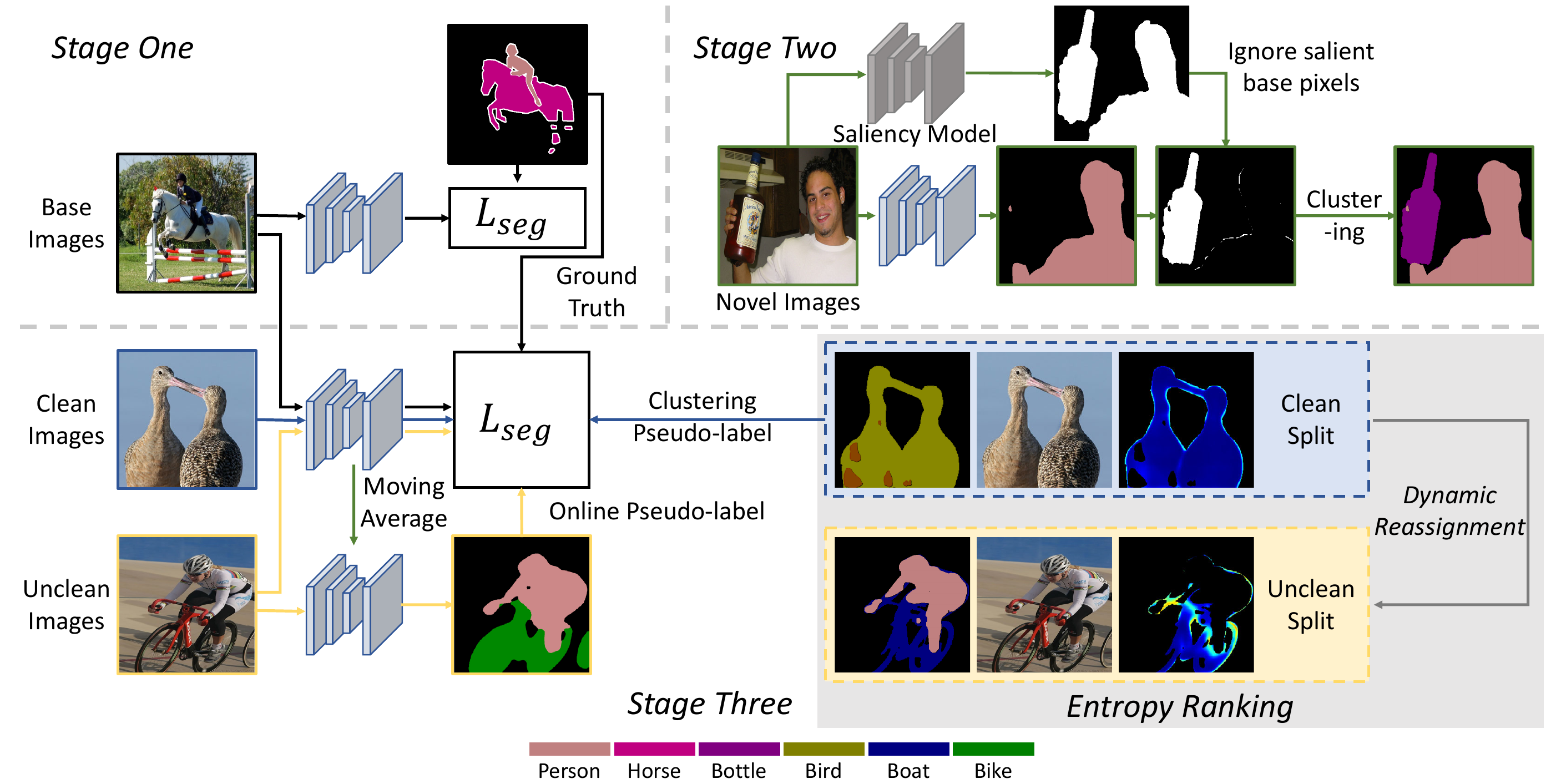}
    \vspace{-.1in}
    \caption{The overview of Entropy-based Uncertainty Modeling and  Self-training framework. 1) The model is first trained on the labeled base set. 2) For the novel images, the saliency maps are generated from a saliency model and high confidence base pixels are ignored. We then apply clustering to the novel salient part to obtain clustering pseudo-labels. 3) We apply entropy ranking and dynamic reassignment to split clean and unclean novel images
    and utilize them to optimize the model with clustering labels and online labels, respectively.}
    \label{fig:framework}
\end{figure*}
%

\paragraph{Semantic Segmentation with Incomplete Labels.}
Such settings are widely explored to alleviate the annotation cost, including but not limited to unsupervised domain adaptation (UDA), semi-supervised learning (SSL), unsupervised learning~(USL) and weak-shot semantic segmentation (WSHSS).
UDA~\cite{adaptseg,advent,zhao2021source} trains a model with labeled source data and unlabeled target data, which focuses on narrowing the domain shift between source and target domains (\eg, synthetic \vs real).
SSL~\cite{zou2021pseudoseg, chen2021semi} concentrates on efficiently utilizing scarce labeled and enormous unlabeled data of the same label space. 
USL~\cite{maskcontrast,cho2021picie} aims to segment different categories by only the unlabeled data and therefore getting rid of heavy annotation cost.
WSHSS~\cite{zhou2021weak} focuses on recognizing novel classes, training a model with labeled base data and weakly labeled data (image-level label) of novel classes. Our NCDSS is more difficult but more practical than WSHSS, where novel data are fully unlabeled.

\vspace{-0.18in}\paragraph{Novel Class Discovery (NCD).}
The concept of clustering novel categories based on prior knowledge can be traced back to Hsu \etal~\cite{hsu2017learning}. Hse \etal~\cite{hsu2017learning} learn the pair-wise image similarities with labeled data, and use such knowledge as the supervision for unsupervised clustering model. Recently, Han \etal~\cite{Han2019learning} formalize the setting and address it in two steps: learning semantic representations from labeled data and fine-tuning the model with unlabeled data clustering. Based on the two-stage pipeline, Han \etal~\cite{han2020rankstat} introduce self-supervised learning to bootstrap feature representation and ranking statistics to explore similarities within unlabeled samples. Zhao \etal~\cite{zhao21novel} further improve the pipeline by dual ranking statistics and mutual knowledge distillation. In addition to better transferring knowledge, another mainstream~\cite{zhong2021openmix, zhong2021neighborhood, Fini2021uod} is to address NCD by exploring noisy pseudo-labels with contrastive learning and labeled base data. In contrast to the above-mentioned methods, we are first to introduce NCD to semantic segmentation. Our setting is much more complex but practical, where more than one categories exist in a single image.

\vspace{-0.18in}\paragraph{Uncertainty Modeling.}
The effectiveness of uncertainty modeling has been demonstrated in unsupervised domain adaptive semantic segmentation~\cite{zheng2021rectifying, advent, intra}. Entropy measurement is one of the key approaches to address uncertainty, which plays an important role in utilizing unlabeled target domain in UDA~\cite{liang2020we, long2016unsupervised, advent}. Pan \etal~\cite{intra} further utilize entropy-based ranking to split unlabeled target data into different parts and use them to narrow intra-domain gap. The aim of entropy ranking in Pan~\etal~\cite{intra} is to find an intermediate domain (the easy part) within the target domain and to improve the hard part learning that cannot be directly learned by inter-domain adaptation.
Despite being inspired by \cite{intra}, %
our entropy ranking aims to distill the more trustworthy clustering pseudo-labels to reduce noisy data that can impair the novel class representations.

\section{Our Methodology}

\paragraph{Problem Definition.}
In NCDSS, we have a labeled base set $D^l = \left\{\mathcal{X}^l, \mathcal{Y}^l \right\}$ containing $N_C^l$ categories $C^l$ and an unlabeled novel set $D^u = \left\{\mathcal{X}^u \right\}$ containing $N_C^u$ new categories $C^u$.
$C^l$ and $C^u$ satisfy $C^l \cap C^u = \emptyset $ and $ C^l \cup C^u = C$. Following \cite{zhong2021neighborhood,han2020rankstat}, the number of novel classes $N_C^u$ is a known prior. The background class is viewed as the base class. 
The goal of NCDSS is to segment images containing 
novel classes $C^u$ by leveraging both $D^l$ and $D^u$.

\vspace{-0.18in}\paragraph{Overview.}
In this section, we first propose a basic framework and then introduce the modifications to better utilize novel data. The overall framework is shown in Fig.~\ref{fig:framework}, which contains three stages, including base training, clustering pseudo-labeling and novel fine-tuning stage.
In the base training stage, we first train the model with the labeled base data. The base model is then used to filter out salient base pixels and assign base labels in the novel images. In the clustering pseudo-labeling stage, novel images are fed into the saliency model and base model to obtain novel foreground pixels. Subsequently, the mean features of novel pixels are used for clustering and assigning novel labels.
For the basic framework, the generated clustering pseudo-labels are used to train the model with segmentation loss. 
To address the noisy clustering pseudo-labels in the basic framework, we propose the Entropy-based Uncertainty Modeling and Self-training (EUMS) framework to improve the novel fine-tuning stage. EUMS dynamically splits and reassigns novel data into clean and unclean parts based on entropy ranking. 
We use the clustering pseudo-labels for the clean part, while online pseudo-labels from a teacher model are used to supervise the unclean data. 
The training procedure is shown in Alg.~\ref{alg:overall}. In the following subsections, we introduce the three stages in detail.

\subsection{Base Training}
The model $f_b$ is first trained on the labeled base data $D^l$ with the standard cross-entropy loss:
\begin{equation}
\label{eq:base}
    \mathcal{L}_{base} = - \frac{1}{HW} \sum\limits_{h=1}^{H} \sum\limits_{w=1}^{W} y^l_{(h,w)} \log f_b(x^l)_{(h,w)},
\end{equation}
where $y^l_{(h,w)}$ denotes the ground truth for one pixel and $f_b(x^l)_{(h,w)}$ denotes the class-wise probabilities of this pixel. Note that it is inevitable that there are images %
containing both novel and base categories, and we view the novel categories as the background in this stage.

\subsection{Clustering Pseudo-labeling}
Directly clustering global features like image classification introduces high noise since the background class and multiple foreground classes exist in one image simultaneously. 
To mitigate this problem, we utilize the base model $f_b$ and a well-trained saliency model $g$ to obtain the salient novel maps for the unlabeled novel images in $D^u$. 
Both base and novel classes are unlabeled in $D^u$. Specifically, the novel images are put into the saliency model to obtain saliency maps $m^u = g(x^u)$ and fed into the base model $f_b$ to obtain high confidence one-hot base pseudo-labels:
\begin{equation}
\label{eq:base-removal}
    \hat{y}^u_b = [\mathbbm{1} _{\max f_b(x^u)_{(h,w,c)}>\tau}] \argmax_c f_b(x^u)_{(h,w,c)},
\end{equation}
where $\tau$ denotes the threshold for selecting high confidence base pixels.
The salient novel maps can then be obtained by ignoring salient base pixels in the saliency maps:
\begin{equation}
\label{eq:novel-map}
    m^u_n = m^u \odot (\mathbbm{1}_{\hat{y}^u_b = 0}),
\end{equation}
where $\mathbbm{1}_{\hat{y}^u_b = 0}$ denotes the inconspicuous base pixels in the image $x^u$, and $\odot$ denotes the Hadamard product. Subsequently, the mean features of the salient novel pixels are used for K-Means Clustering~\cite{macqueen1967some_kmeans} to obtain image-level clustering labels. The final clustering segmentation pseudo-labels $\hat{y}^u$ are the fusion of base pseudo-labels and clustering novel pseudo-labels:
\begin{equation}
\label{eq:clustering-labels}
    \hat{y}^u = \hat{y}^u_b + c^u \cdot m^u_n,
\end{equation}
where $c^u$ denotes the image-level clustering labels.

\begin{figure}[t]
    \centering
    \includegraphics[width=0.9\linewidth]{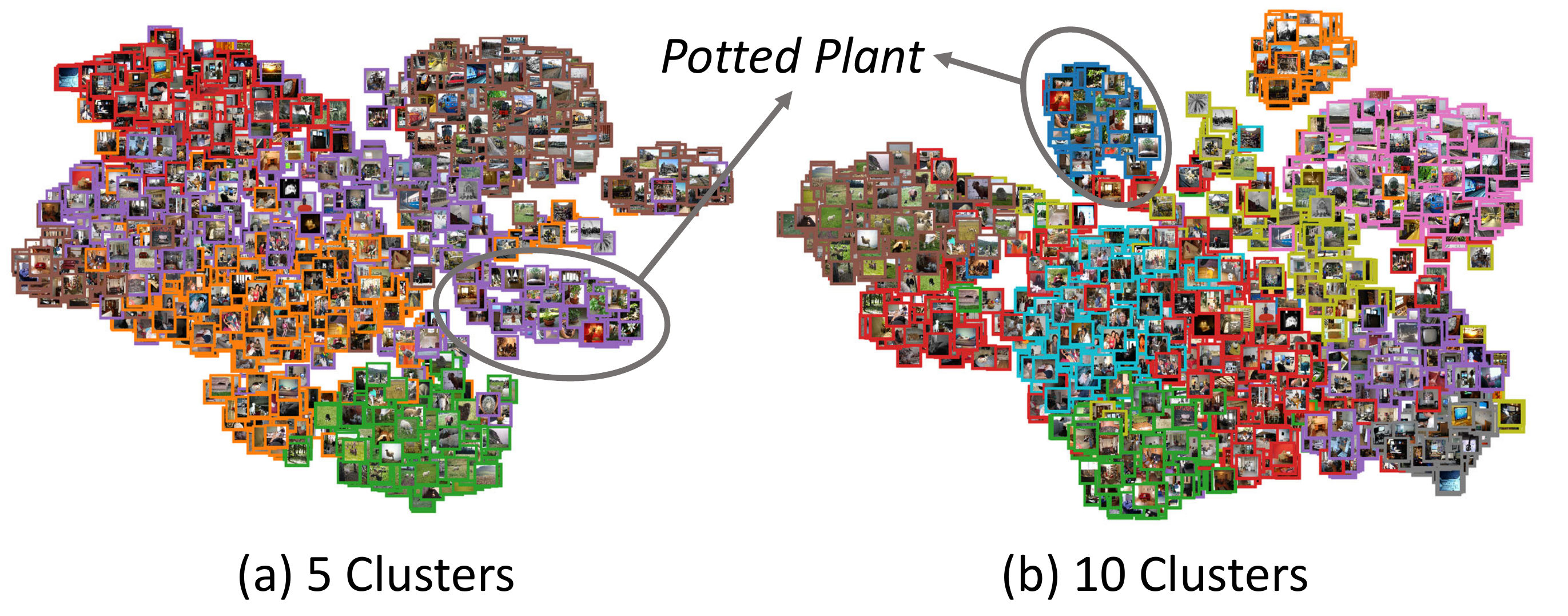}
    \vspace{-.1in}
    \caption{T-SNE~\cite{tsne} visualization of different number of clusters on PASCAL-5$^i$ Fold3. The exact novel class number $N_C^u$ is 5. Zoom in for details.}
    \label{fig:overclustering}
\end{figure}

\vspace{-0.18in}\paragraph{Over-Clustering.} Since we know the number of novel classes $N_C^u$ (\eg, 5 for PASCAL-5$^i$ dataset), the intuitive way is to cluster the novel images into $N_C^u$ clusters. However, hard classes can be easily ignored when clustering into exact clusters
due to the inaccurate saliency maps and multiple novel classes existing in one image. 
An example is shown in Fig.~\ref{fig:overclustering}. The potted plant images are mixed with other images into the purple cluster when only 5 clusters exist (Fig.~\ref{fig:overclustering}(a)). Inspired by \cite{overclustering}, we adopt over-clustering to improve the clustering of hard classes. In Fig.~\ref{fig:overclustering}(b), the potted plant images can be well clustered into the blue cluster when double clusters are used.

\subsection{Novel Fine-tuning}

\subsubsection{Basic Framework}

We can now train a segmentation model $f$ to discover novel classes with the labeled base data and pseudo-labeled novel data. Our basic framework uses the standard segmentation loss (cross-entropy loss) as supervision:
\begin{equation}
\label{eq:basic}
    \mathcal{L}_{basic} = \mathcal{L}_{seg}(x^l, y^l; f) + \mathcal{L}_{seg}(x^u, \hat{y}^u; f),
\end{equation}
where $\hat{y}^u$ denotes the final clustering segmentation pseudo-labels, which can be obtained by either exact-clustering or over-clustering.

\subsubsection{EUMS Framework}
\paragraph{Entropy Ranking.}
Despite over-clustering, the pseudo-label accuracy is still limited by the saliency maps and the randomness of clustering. In addition, multiple novel classes may exist in one novel image while clustering can only assign one label for one image. Taking all the above into consideration, clustering pseudo-labels are still noisy and can impair the model performance~\cite{arpit2017closer, zhang2017understanding}. 
Inspired by \cite{advent, intra}, the uncertainty of novel images can be represented by the entropy, and the uncertainty of novel foreground pixels from basic model reflects the accuracy of pseudo-labels.
Thus, we use entropy-based uncertainty modeling to select the clean labels and discard the unclean ones. Specifically, the image entropy is formulated as:
\begin{equation}
\label{eq:ent-map}
{E}^u_{(h, w)}= - \frac{1}{\log (C)} \sum_{c=1}^{C} {P}^u_{(h, w, c)} \log {P}^u_{(h, w, c)},
\end{equation}
where ${P}^u_{(h, w, c)}=f(x^u)_{(h,w,c)}$ denotes the probability of pixel at position $(h,w)$ belonging to the $c$-th class. $C=C^l+C^u$ is the number of all categories. The foreground entropy is the average entropy of novel salient pixels:
\begin{equation}
\label{eq:ent-value}
    \overline{E}^u = \frac{1}{\sum \mathbbm{1}_{m^u_n} } \sum_{h,w \in \mathbbm{1}_{m^u_n}} m^u_n \odot E^u_{(h, w)},
\end{equation}
where $m^u_n$ denotes the novel saliency maps, and $\mathbbm{1}_{m^u_n}$ denotes the detected novel pixels. With the foreground entropy of the novel images, we split the lower $\lambda$ ratio of images as clean data and the others as unclean data.

\begin{algorithm}[t]
    \caption{The training procedure of EUMS framework.}
    \label{alg:overall}
    \textbf{Inputs:} labeled base set $D^l$, unlabeled novel set $D^u$, base segmentation model $f_b$, novel segmentation model $f$, and saliency model $g$. \\
    \textbf{Outputs:} Optimized model $f$ for segmenting novel classes.\\
    \vspace{-.15in}
    \begin{algorithmic}[1]
        \STATE  \textcolor{gray}{// Stage 1: Base Training.}
        \FOR{$i$ in $Base Epochs$}
  \STATE Sample mini-batch $x^l$ from base set $D^l$;
  \STATE Optimize the base model $f_b$ with Eq.~\ref{eq:base};
        \ENDFOR
        \STATE  \textcolor{gray}{// Stage 2: Clustering Pseudo-labeling.}
        \FOR{novel images $x^u$ in $D^u$}
  \STATE Obtain saliency maps $m^u$ from saliency model $g$;
  \STATE Ignore salient base pixels with Eq.~\ref{eq:base-removal} and Eq.~\ref{eq:novel-map};
  \STATE Assign clustering labels by K-Means Clustering;
  \STATE Fuse pseudo-labels with Eq.~\ref{eq:clustering-labels};
        \ENDFOR
        \STATE  \textcolor{gray}{// Stage 3: Novel Fine-tuning.}
        \FOR{$j$ in $Novel Epochs$}
  \STATE Sample mini-batch $x^l$ from base set $D^l$;
  \STATE Sample mini-batch $x^u$ from novel set $D^u$;
  \STATE Optimize the basic model $f$ with Eq.~\ref{eq:basic};
        \ENDFOR
        \STATE Split clean and unclean data with entropy ranking Eq.~\ref{eq:ent-value};
        
        \FOR{$k$ in $Novel Epochs$}
  \IF{$k = 5$}
  \STATE Dynamically reassign clean part with Eq.~\ref{eq:ent-value};
  \ENDIF
  \STATE Sample mini-batch $x^l$ from base set $D^l$;
  \STATE Sample mini-batch $x^u_c$ from clean part $D^u$;
  \STATE Sample mini-batch $x^u_d$ from unclean part $D^l$;
  \STATE Generate online pseudo-labels with Eq.~\ref{eq:online-label};
  \STATE Optimize the basic model $f$ with overall loss Eq.~\ref{eq:overall-loss};
        \ENDFOR   
        
        \STATE \textbf{Return} Novel segmentation model $f$.
    \end{algorithmic}
\end{algorithm}

\vspace{-0.18in}\paragraph{Dynamic Reassignment.}
Intuitively, selecting a smaller ratio $\lambda$ can distill more accurate pseudo-labels and help to learn a better model with other techniques like self-training. However, hard classes (\eg, potted plant, sofa and chair) are almost ignored when $\lambda$ is too small (\eg, 0.33) and this leads to the model being unable to detect such classes. Arpit \etal~\cite{arpit2017closer} have made the observation that deep neural networks can memorize easy instances first, and then gradually fit hard samples along with training. 
This observation means that we can re-rank the entropy of the clean split and distill cleaner pseudo-labels to further improve the performance. Specifically, we reassign half of the clean data to the unclean split after 5 epochs, and the unclean pseudo-labels are discarded to avoid degenerating model performance. 
Experiments in Sec.~\ref{sec:ablation} demonstrate that the model can achieve comparable performance to the basic model only with clean images from entropy ranking, and can achieve further improvement with dynamic reassignment.

\begin{table*}[t]
\begin{center}
\begin{tabular}{cccc|ccccc}
\toprule
Over- & Entropy & Dynamic & Self- & \multicolumn{5}{c}{PASCAL-5$^i$} \\ 
Clustering & Ranking & Reassignment & Training &  {Fold0} & {Fold1} & {Fold2} & {Fold3} & AVG \\
\midrule
-- & -- & -- & -- & {60.83} & {58.90} & {42.53} & {36.96} & 49.81 \\  
\cmark& -- & -- & -- & {62.43} & {57.02} & {51.53} & {47.05} & 54.51 \\  
\cmark& \cmark& -- & -- & {63.22} & {56.26} & {49.82} & {48.12} & 54.36 \\  
\cmark& \cmark& \cmark& -- & {61.36} & {57.90} & {55.10} & {46.80} & 55.29 \\  
\cmark& \cmark& -- & \cmark& {65.12} & {57.49} & {\textbf{57.46}} & {46.55}  & 56.66 \\ 
\cmark& \cmark& \cmark& \cmark& {\textbf{69.79}} & {\textbf{60.11}} & {56.28} & {\textbf{50.18}} & \textbf{59.09} \\ 
\bottomrule
\end{tabular}
\end{center}
\vspace{-.2in}
\caption{Ablation studies of different components in EUMS framework. The method without any components is our basic framework.}
\label{tab:ablation}
\end{table*}

\vspace{-0.18in}\paragraph{Self-Training.}
Although the model can outperform the basic model only with clean data, the unclean data without clustering pseudo-labels can also further boost the performance. For the unclean data, generated online pseudo-labels can be much more accurate than the clustering labels, which can help the model training. Specifically, following \cite{sohn2020fixmatch}, we adopt a teacher model to generate pseudo-labels $\hat{y}^u_{d}$ for images of weak augmentations. 
\begin{equation}
\small
\label{eq:online-label}
    \hat{y}^u_{d} = [\mathbbm{1} _{\max \overline{f}(\alpha(x^u_{d}))_{(h,w,c)}>\eta}] \argmax_c \overline{f}(\alpha(x^u_{d}))_{(h,w,c)},
\end{equation}
where $\alpha(x^u_{d})$ denotes the weakly augmented images of unclean split, and $\eta$ denotes the threshold for selecting trustworthy pixels. $\overline{f}$ denotes the teacher model updated by moving average.
The online pseudo-labels are used to supervise the student model with strongly augmented images:
\begin{equation}
\label{eq:unclean}
    \mathcal{L}_{d} = - \frac{1}{HW} \sum\limits_{h=1}^{H} \sum\limits_{w=1}^{W} \hat{y}^u_{d (h,w)} \log f(\beta(x^u_{d}))_{(h,w)},
\end{equation}
where $\beta(x^u_{d})$ denotes unclean split images with strong augmentation.

\vspace{-0.18in}\paragraph{Overall Loss Function.}
The overall loss function is the combination of the base data segmentation loss, clean split segmentation loss and unclean split self-supervised loss:
\begin{equation}
\label{eq:overall-loss}
\begin{aligned}
    \mathcal{L}_{overall} = \mathcal{L}_{seg}(x^l, y^l; f) &+ \mathcal{L}_{seg}(x^u_c, \hat{y}^u_c; f) \\
    &+ \omega (t) \mathcal{L}_{d}(x^u_d, \hat{y}^u_d; f),    
\end{aligned}
\end{equation}
where $x^l$, $x^u_c$ and $x^u_d$ denote base, clean split and unclean split images, respectively. $y^l$, $\hat{y}^u_c$ and $\hat{y}^u_d$ denote ground truth labels, clustering pseudo-labels and online pseudo-labels, respectively. Coefficient $\omega (t)$ is a ramp-up function. Following~\cite{laine2017temporal, tarvainen2017mean}, we use the sigmoid-shaped function $\omega(t)= e^{-5\left(1-\frac{t}{T}\right)^{2}}$, where $t$ is current epoch and $T$ is the ramp-up length. We set $T$ as 5 in all experiments.

\section{Experiments}

\subsection{Experimental Setup}
\paragraph{Dataset.}
We conduct experiments on the PASCAL-5$^i$ dataset~\cite{pascal5i} and the COCO-20$^i$ dataset~\cite{coco20i}. PASCAL-5$^i$ is built based on PASCAL VOC 2012 dataset~\cite{voc}, which contains 21 classes, including 20 foreground object classes and one background class. 
Following previous works~\cite{maskcontrast, zhou2021weak}, we use the extended training set~\cite{trainaug} (10,582 images) as our training set and evaluate the model on the validation set (1,449 images).
We split the 20 foreground object categories into 4 folds. For each fold, the split 5 classes are viewed as novel classes while the remaining 15 foreground classes and the background class are viewed as base classes.
COCO-20$^i$ is modified from MS COCO 2014~\cite{coco}, which contains 82,081 training images and 40,137 validation images of 80 foreground and one background classes. 
Following~\cite{coco20i}, we split the 80 foreground classes into 4 folds, and thus each fold contains 20 different classes, which are viewed as novel classes.
In the base training stage, all images containing base classes are used as training samples, and the novel classes here are viewed as the background class. In the clustering pseudo-labeling and novel fine-tuning stage, the full training set is used, where all images containing novel classes are unlabeled.

\vspace{-0.18in}\paragraph{Implementation Details.}
We use DeepLab-v3~\cite{deeplabv3} with ResNet-50 backbone~\cite{resnet} as our segmentation model. For the base training stage, the model is trained for 60 epochs with a batch size of 16. SGD with momentum 0.9 and weight decay 0.0001 is used for optimization. 
The learning rate is initialized to 0.1 for the decoder and 0.001 for the layer 3 and 4 of the backbone, and then it is decayed by 0.1 at the 25th epoch. Images are resized to 512$\times$512 and random flipping, scale and rotation are applied as data augmentation. 
In the clustering pseudo-labeling stage, following~\cite{maskcontrast}, we adopt BAS-Net~\cite{qin2019basnet} trained on DUTS~\cite{duts} to estimate saliency maps of all the novel images. The threshold $\tau$ is set to 0.9 to obtain high confidence base pixels. K-Means clustering is applied on the features from the layer 4 of ImageNet~\cite{imagenet} pre-trained ResNet-50 backbone, and the cluster number is set to 5 and 10 for exact-clustering and over-clustering respectively.
In the novel fine-tuning stage, the model is trained for 30 epochs with a batch size of 8 for all of base, clean and unclean parts. We use the same optimizer and initial learning rate as the base training stage, and the learning rate is decayed by 0.1 at the 15th epoch. We set the clean part ratio $\lambda$ as 0.67. For self-supervised learning, we use the augmentation in base training as the weak augmentation $\alpha$, and color jitter is applied as strong augmentation $\beta$. The threshold of online pseudo-labeling $\eta$ is set to 0. We conduct parameter analysis on the clean part ratio $\lambda$ and online pseudo-labeling threshold $\eta$ in Sec.~\ref{sec:param}. All models are trained with one GTX 3090 GPU (24 GB).

\vspace{-0.18in}\paragraph{Evaluation Metric.}
Experiments are conducted on all four folds, and mean Intersection-over-Union (mIoU) of novel classes is adopted as the evaluation metric.

\subsection{Ablation Studies}
\label{sec:ablation}

To investigate the effectiveness of the basic framework and different components of EUMS in the novel fine-tuning stage, we conduct ablation studies on PASCAL-5$^i$ in Tab.~\ref{tab:ablation}. 

\vspace{-0.18in}\paragraph{Baseline.} The baseline is the basic framework, which uses clustering pseudo-labels as supervision (Eq.~\ref{eq:basic}). As shown in Tab.~\ref{tab:ablation}, the baseline achieves an average mIoU of 49.81\% on the four folds. The performance of Fold2 and Fold3 is poorer since the saliency maps for hard classes (\eg, dining table, potted plant, and sofa) are not accurate and several novel classes exist in one image (\eg, horse and person). The results suggest the feasibility and limitation of the basic framework. 

\vspace{-0.18in}\paragraph{Effectiveness of Over-Clustering.}
The second row of Tab.~\ref{tab:ablation} demonstrates the effectiveness of over-clustering in generating more accurate pseudo-labels. Over-clustering obtains 4.7\% improvement in average mIoU from the baseline. 
The improvement of Fold2 (9.00\%) and Fold3 (10.09\%) is more than that of Fold0 (1.6\%) and Fold1 (-1.88\%). This is because over-clustering is good at discovering hard classes but may introduce unpredictable noise for some easier classes. For example, the ``cat'' class in Fold1 are clustered into several clusters when utilizing over-clustering, but some of those clusters contain both ``cat'' and ``cow''. Those confusing clusters degrade the performance on both of the two categories. 
Without over-clustering, ``cat'' and ``cow'' are clustered into two clusters, respectively, and no noisy clusters are introduced.

\vspace{-0.18in}\paragraph{Effectiveness of Entropy Ranking and Dynamic Reassignment.}
Entropy ranking splits the clustering pseudo-labels into clean and unclean parts, which is the basis of EUMS. In Tab.~\ref{tab:ablation}, applying only entropy ranking (third row) means training the segmentation model only with the clean part ($\lambda$ ratio of data with lower entropy). Despite using only a portion of the data, the model can achieve comparable performance with the basic framework (54.36\% \vs 54.51\% average mIoU). Additionally, dynamically distilling clean pseudo-labels can further improve the performance. Without other techniques, utilizing entropy ranking and dynamic reassignment gains 0.78\% improvement in average mIoU.

\begin{figure}[t]
    \centering
    \includegraphics[width=.95\linewidth]{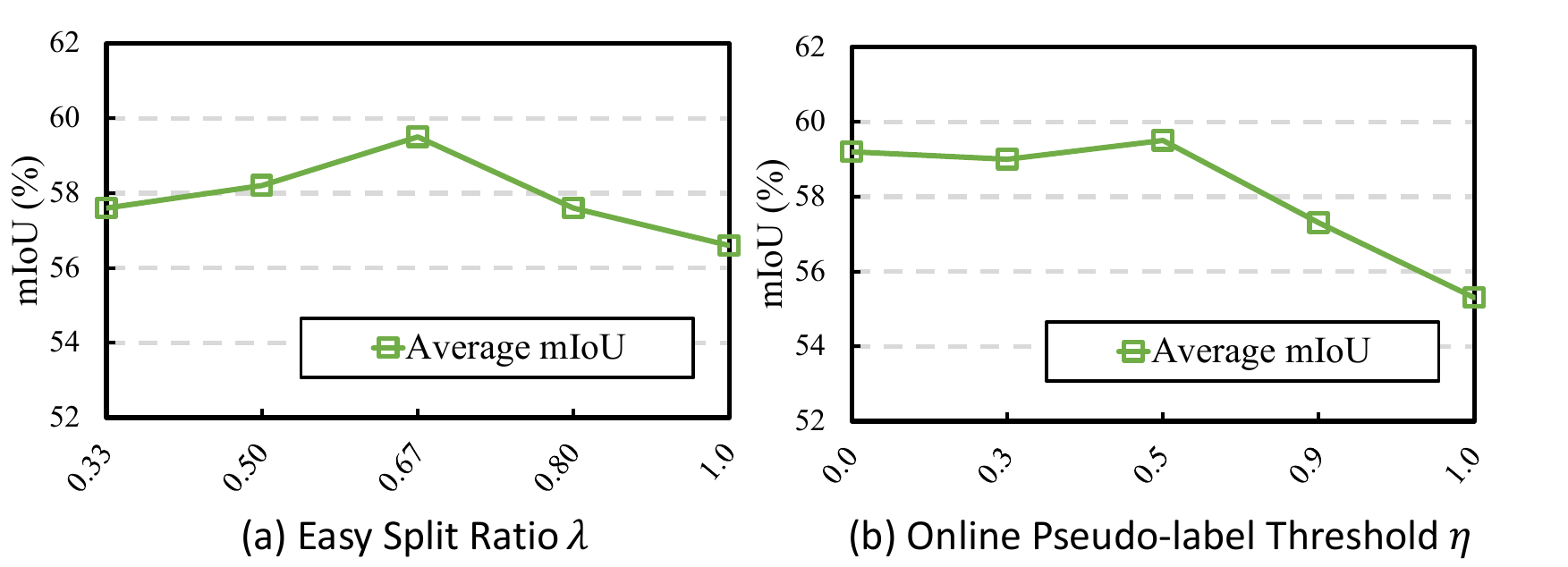}
    \vspace{-.1in}
    \caption{Parameter analysis of easy split ratio $\lambda$ and online pseudo-label threshold $\eta$.}
    \vspace{-.1in}
    \label{fig:param}
\end{figure}

\vspace{-0.18in}\paragraph{Effectiveness of Self-Training.}
Our method with only clean data can already achieve competitive results with the basic framework.
In Tab.~\ref{tab:ablation}, we show that the performance of our model can be further improved by self-training.
First, we disentangle the influence of self-training in the fifth row of Tab.~\ref{tab:ablation}. 
Utilizing the unclean data by self-training yields 2.3\% improvement in average mIoU when the clean and hard parts are split at the beginning of training (without dynamic reassignment). 
Second, the segmentation model achieves an average mIoU of 59.09\%, outperforming the basic framework by 9.28\% when over-clustering is combined with all components in EUMS. 
All results in Tab.~\ref{tab:ablation} demonstrate that EUMS can make full use of unlabeled data to achieve better performance by uncertainty modeling and self-training.

\subsection{Parameter Analysis}
\label{sec:param}

We further analyze the sensitivity of EUMS to two important hyper-parameters on PASCAL-5$^i$ dataset, \ie, the easy split ratio $\lambda$ of entropy ranking and the online pseudo-label threshold $\eta$ of self-training.

\vspace{-0.18in}\paragraph{Easy split ratio $\lambda$.} 
$\lambda$ plays an important role in learning a clean but complete initial knowledge in the novel fine-tuning stage. We compare the results of using different easy split ratio $\lambda$ in Fig.~\ref{fig:param}(a). 
Many pseudo-labels are discarded when $\lambda$ is set to a small value (0.33 and 0.50), especially those from the hard classes, which leads the model to learn incomplete knowledge. The average mIoU increases and peaks when $\lambda=0.67$. The performance gradually degrades when $\lambda$ increases from 0.67 to 1.0. This is because many inaccurate pseudo-labels are included into the clean part and thus introducing enormous noise.  

\vspace{-0.18in}\paragraph{Online pseudo-label threshold $\eta$.}
The quality of online pseudo-labels is sensitive to the threshold $\eta$. 
Previous works~\cite{sohn2020fixmatch, zhang2021flexmatch} commonly use fixed weight (\eg, 1) for the unlabeled loss term and set a high threshold (\eg, $>0.9$) to filter out the low probability labels since the model is less confident at the beginning epochs. Nevertheless, we use a ramp-up weight for the unclean data loss to gradually increase the impact of self-training. 
We investigate the effect of $\eta$ in Fig.~\ref{fig:param}(b) and make distinct observations. The performance is impacted just marginally when $\eta$ is within a small scale ($\eta < 0.5$), but drops by a considerable margin of 3\% when $\eta = 0.9$. $\eta=1.0$ denotes removing the self-training. We conjecture that this is because ramp-up weight can reduce the impact of noisy labels in the early training stage and make better use of unlabeled data in the later stage with relatively small $\eta$. Additional useful information is lost when $\eta$ is high. We set $\eta=0$ in our framework.

\begin{table}[t]
\small
\begin{center}
\begin{tabular}{c|ccccc}
\toprule
\multirow{2}{*}{Method} &\multicolumn{5}{c}{COCO-20$^i$} \\ 
&{Fold0} & {Fold1} & {Fold2} & {Fold3} & AVG \\
\midrule
Basic & 24.58 & 24.14 & 15.36 & 15.57 & 19.91 \\
Basic-OC & 39.39 & 25.01 & 17.95 & 15.53 & 24.47 \\
EUMS & \textbf{42.39} & \textbf{26.89} & \textbf{19.75} & \textbf{18.19} & \textbf{26.81} \\
\bottomrule
\end{tabular}
\end{center}
\vspace{-.2in}
\caption{Performance on COCO-20$^i$ benchmark. %
}
\vspace{-.1in}
\label{tab:coco}
\end{table}

\begin{figure*}[t]
    \centering
    \includegraphics[width=\linewidth]{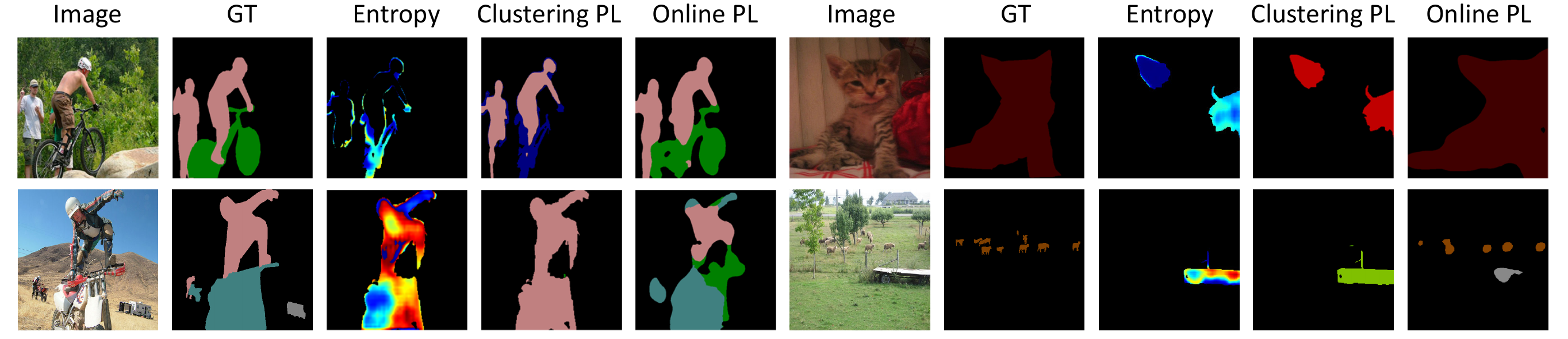}
    \vspace{-.3in}
    \caption{Visualization of entropy maps and pseudo-labels in PASCAL-5$^i$ dataset. ``Entropy'', ``Clustering PL'', and ``Online PL'' denote the entropy maps of the salient novel pixels, clustering pseudo-labels and online pseudo-labels. The four images (left to right, top to bottom) contain novel classes of Fold0 to Fold3, respectively.}
    \label{fig:visual-labels}
\end{figure*}

\begin{figure*}[t]
    \centering
    \vspace{-.1in}
    \includegraphics[width=\linewidth]{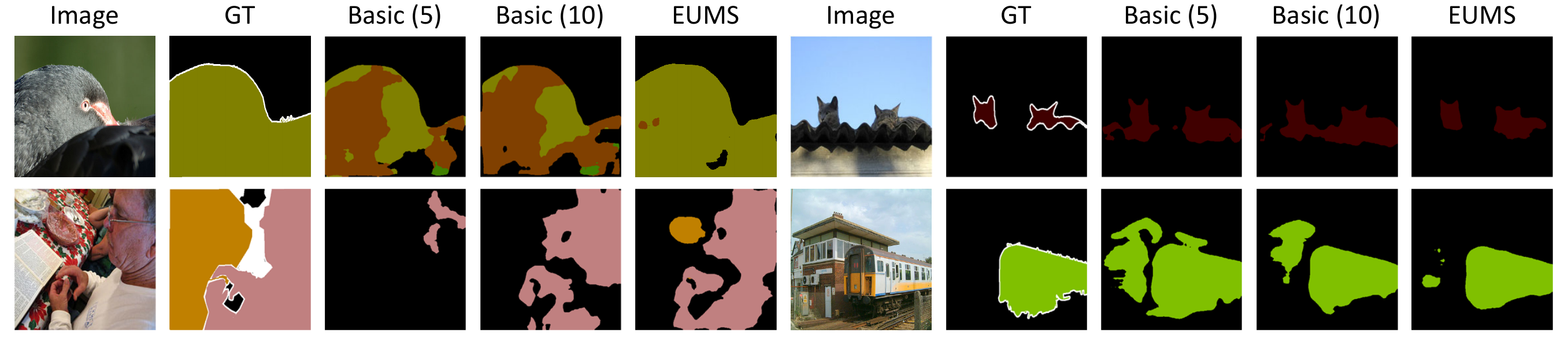}
    \vspace{-.3in}
    \caption{Qualitative comparison of segmentation results in PASCAL VOC 2012 validation set. ``Basic (5)'' and ``Basic (10)'' denote the basic framework with 5 and 10 clusters. The four images (left to right, top to bottom) contain novel classes of Fold0 to Fold3, respectively.}
    \vspace{-.15in}
    \label{fig:visual-results}
\end{figure*}

\subsection{Performance on COCO-20$^i$ Benchmark}
Compared with PASCAL-5$^i$~\cite{pascal5i}, COCO-20$^i$~\cite{coco20i} is a much more challenging benchmark with more novel classes and more data.
We compare the basic framework (Basic), basic framework with over-clustering (Basic-OC), and the EUMS framework (EUMS) in Tab.~\ref{tab:coco}. We make two observations. First, applying over-clustering can yield significant improvement on Fold0 (14.81\%) and Fold2 (2.59\%) while making little progress on Fold1 and Fold3. This mainly lies in that there exist many similar and correlated object classes in Fold0 and Fold2 (\eg, bicycle, motorcycle and person), which can be better separated by over-clustering.
Second, the proposed EUMS framework consistently improves the performance on all of the four folds, 
which outperforms the baseline by 6.9\% in average mIoU.

\subsection{Visualization}

\paragraph{Visualization of Pseudo-labels.}
We illustrate the entropy maps and pseudo-labels of unclean data in Fig.~\ref{fig:visual-labels} to demonstrate the effectiveness of EUMS in addressing noisy clustering pseudo-labels. As shown in Fig.~\ref{fig:visual-labels}, the entropy values of salient novel pixels in unclean data are high. In addition, the clustering pseudo-labels of high entropy data tend to be noisy due to imprecise clustering~(first image), inaccurate saliency maps (second and fourth images), and the simultaneous existence of multiple novel categories (third image). Our EUMS discards the noisy clustering pseudo-labels of unclean data and utilizes those data in a self-supervised manner by online pseudo-labels. The generated online pseudo-labels are much better than those from clustering, which can further improve the performance.

\vspace{-0.18in}\paragraph{Visualization of Segmentation Results.}
We compare the segmentation results of EUMS with those of the basic framework in Fig.~\ref{fig:visual-results}. ``Basic (5)'' and ``Basic (10)'' denote the basic framework with exact-clustering and over-clustering respectively. As shown in Fig.~\ref{fig:visual-results}, over-clustering can significantly boost the performance in some folds (Fold2 and Fold3), but achieves similar or even worse results in the others (Fold0 and Fold1). Our EUMS can consistently outperform the baseline on all of the four folds, no matter on easy categories (\eg, bird and train) or hard categories (\eg, dining table).

\section{Conclusion}
In this work, we introduce a new setting of Novel Class Discovery in Semantic Segmentation (NCDSS). The proposed setting aims at segmenting unlabeled novel classes with prior knowledge from labeled data of disjoint categories. %
To address this problem, we propose a basic framework, leveraging labeled base data and a saliency model to discover novel classes in unlabeled images by clustering. In addition, we propose the Entropy-based Uncertainty Modeling and Self-training (EUMS) framework to overcome the noisy labels in the basic framework for further improvements. Extensive experiments on the built benchmark of PASCAL-5$^i$ and COCO-20$^i$ demonstrate the feasibility of the basic framework and the effectiveness of EUMS.

\section*{Acknowledgements}
This research is supported in part by the National Research Foundation, Singapore under its AI Singapore Program (AISG Award No: AISG2-RP-2020-016), the Tier 2 grant MOE-T2EP20120-0011 from the Singapore Ministry of Education, the EU H2020 projects SPRING No. 871245, and AI4Media No. 951911. 


\newpage

{\small
\bibliographystyle{ieee_fullname}
\bibliography{egbib}
}

\clearpage

\appendix

\section{Performance of All Classes}
Despite the focus on novel classes as previous works~\cite{zhong2021neighborhood, han2020rankstat}, our NCDSS setting still maintains the ability of segmenting base classes. As shown in Tab.~\ref{tab:all-classes}, our EUMS achieves nearly 70\% base mIoU on Fold0 and Fold3, and more than 60\% base mIoU on Fold1 and Fold2. 
However, our performance is not as competitive as the base model of stage one
despite the use of fully-annotated base data. There are two reasons for this limitation. 
First, the model is required to segment all of base and novel classes together, increasing the task difficulty.
Second, there exist unlabeled base classes in the novel images and the generated pseudo-labels of the base classes are also not completely accurate. Thus, bad cases are introduced into the base classes. 
The above two factors limit the base class performance. How to maintain the high base performance while discovering novel classes in semantic segmentation deserves further explorations in the future.

\begin{table}[h]
\centering
\small
\begin{tabular}{c|ccc}
\toprule
\multirow{2}{*}{Fold} & \multicolumn{3}{c}{mIoU}\\ 
 & {Base}& {Novel}& All \\ 
\midrule
PASCAL-5$^0$ & {69.28}& {69.79}& 69.40\\ 
PASCAL-5$^1$ & {66.95}& {60.11}& 65.32\\ 
PASCAL-5$^2$ & {62.87}& {56.28}& 61.30 \\ 
PASCAL-5$^3$ & 69.83 & 50.18 & 65.15  \\ 
\bottomrule
\end{tabular}
\vspace{-.1in}
\caption{Performance of all classes.}
\vspace{-.1in}
\label{tab:all-classes}
\end{table}

\section{Comparison with Related Settings}
We further compare several methods under related settings in Tab.~\ref{tab:setting-com} on PASCAL-5$^i$.
Our method clearly outperforms the unsupervised learning method on all the folds. Interestingly, our method performs higher on Fold0 when compared with the methods using image/pixel-level labels. However, these weak\&few-shot methods generally give better results on the other folds. Please note that directly comparing our method with these weak\&few-shot methods is not exactly fair.

\begin{table}[ht]
\centering
\resizebox{.95\linewidth}{!}{
\begin{tabular}{l|c|cccc}
\toprule
Method  & Setting   & Fold0 & Fold1 & Fold2 & Fold3 \\
\midrule
PFENet~\cite{tian2020prior} & \multirow{2}{*}{1-Shot} & 61.7  & 69.5  & 55.4  & 56.3  \\
ASGNet~\cite{li2021adaptive}   & & 58.8  & 67.9  & 56.8  & 53.7  \\
\midrule
PFENet~\cite{tian2020prior}  & \multirow{2}{*}{5-Shot} & 63.1  & 70.7  & 55.8  & 57.9  \\
ASGNet~\cite{li2021adaptive}    &  & 63.7  & 70.6  & 64.2  & 57.4  \\
\midrule
CAM+RETAB~\cite{zhou2021weak} & \multirow{2}{*}{Weak-Shot} & 69.2  & \textbf{76.1}  & 72.0  & 58.5 \\
SEAM+RETAB~\cite{zhou2021weak} & & 65.4  & 74.5  & \textbf{73.0}  & \textbf{58.9}  \\
\midrule
MaskContrast~\cite{maskcontrast} & Unsupervised & 55.3 & 38.9 & 35.6 & 37.0 \\
\midrule
EUMS & NCDSS & \textbf{69.8}  & 60.1  & 56.3  & 50.2  \\
\bottomrule
\end{tabular}}
\vspace{-.1in}
\caption{Comparison with peer methods.}
\label{tab:setting-com}
\end{table}

\begin{figure}[t]
    \centering 
    \includegraphics[width=.95\linewidth]{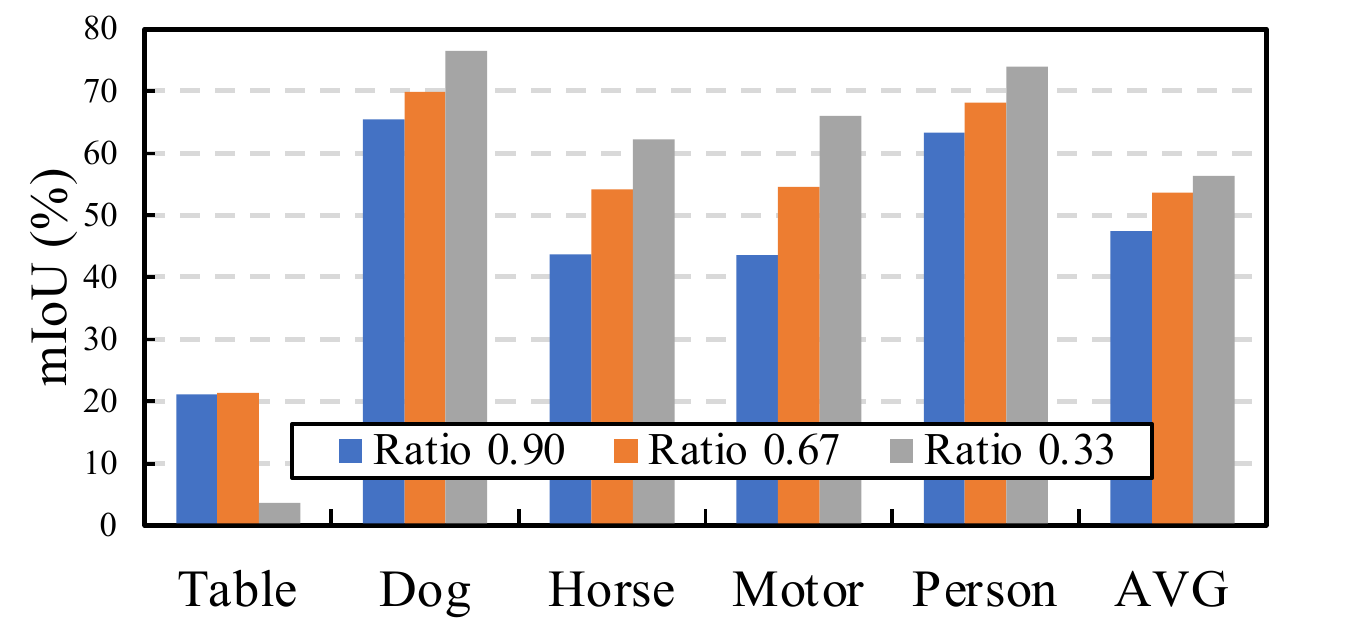} 
    \vspace{-.15in}
    \caption{Accuracy (mIoU) of clustering pseudo-labels with different easy split ratio $\lambda$ in Fold2.}
    \label{fig:pseudo-label}
    \vspace{-.1in}
\end{figure}

\section{Further Explanation on Easy Split Ratio}
 
In our method, we set the easy split ratio $\lambda$ as a hyper-parameter. 
We study its impact on PASCAL-5$^i$ and observe two phenomena. 
First, more incorrect labels are included when $\lambda$ is too large. Second, hard classes will be largely ignored when $\lambda$ is too small. We show the accuracy of pseudo-labels with different $\lambda$ in Fig.~\ref{fig:pseudo-label}. The average accuracy is poor when $\lambda$ is 0.90, while the dinning table is almost ignored when $\lambda$ is 0.33. 
This motivates us to select the easy split ratio and $\lambda=0.67$ is the best choice.

\begin{figure*}[t]
    \centering
    \includegraphics[width=.8\linewidth]{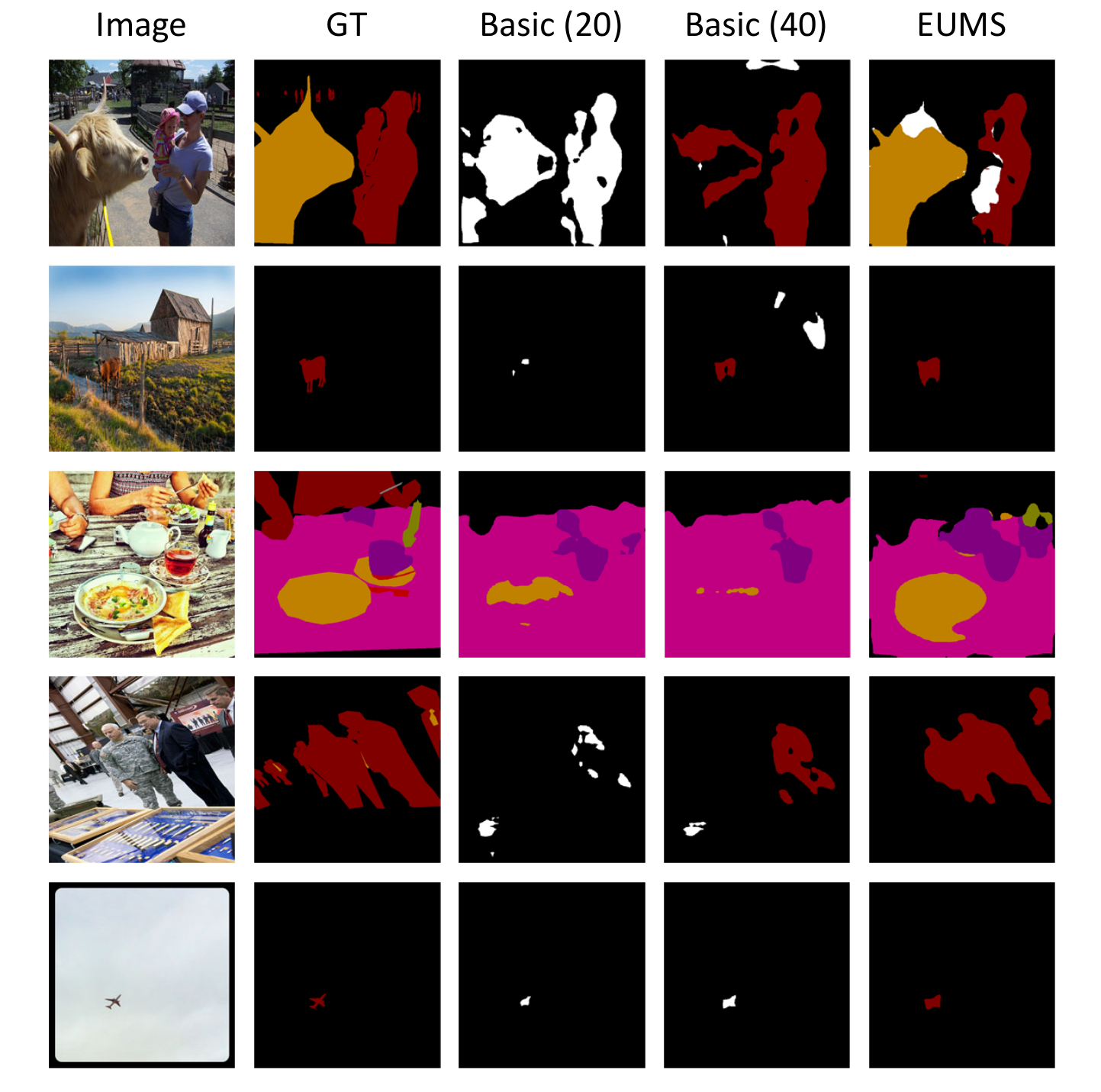}
    \vspace{-.15in}
    \caption{Qualitative comparison of segmentation results in MS COCO 2014 validation set. ``Basic (20)'' and ``Basic (40)'' denote the basic framework with 20 and 40 clusters.}
    \vspace{-.1in}
    \label{fig:coco-visual}
\end{figure*}

\section{Limitation}
Semantic-relevant knowledge between base and novel classes is required for novel class discovery. 
For example, ``potted plant'' in Fold3 is a semantically different class from the base classes. The mIoU of this class is only 34.5\%, which is much less than the other novel classes in Fold3.

\section{Visualization}

We provide the qualitative comparison on COCO-20$^i$~\cite{coco20i} in Fig.~\ref{fig:coco-visual}. Our EUMS can well handle the circumstances that multiple classes exist in one image (the first and third examples) and the cases that the object is tiny and hard to segment (the second and fourth examples).

\end{document}